\documentclass[letterpaper, 10 pt, conference]{ieeeconf}  

\IEEEoverridecommandlockouts                              

\usepackage{cite}
\usepackage{amsmath,amssymb,amsfonts}
\usepackage{bm}
\usepackage{array}
\usepackage{caption}
\usepackage[ruled, linesnumbered]{algorithm2e}
\usepackage{graphicx}
\graphicspath{{./img/}}
\usepackage{textcomp}
\usepackage{xcolor}
\usepackage[utf8]{inputenc}
\usepackage{bm}
\usepackage{flushend}
\usepackage{parskip}
\usepackage{paralist}
\usepackage{graphicx}
\usepackage{xcolor}
\usepackage{acronym}
\usepackage{subcaption}
\usepackage[skip=1pt]{caption}
\usepackage{hyperref}
\usepackage[capitalise]{cleveref}

\def\BibTeX{{\rm B\kern-.05em{\sc i\kern-.025em b}\kern-.08em
    T\kern-.1667em\lower.7ex\hbox{E}\kern-.125emX}}
    
\newif\iftrackchanges

\begin{document}
\trackchangesfalse

\newcommand\X{\ensuremath{\mathcal{X}}}
\newcommand\Xr{\ensuremath{\mathcal{X}_{\textrm{rel}}}}
\newcommand\Xj{\ensuremath{\mathcal{X}_j}}
\newcommand\Q{\ensuremath{\mathcal{Q}}}
\newcommand\Rn{\ensuremath{\mathbb{R}^n}}
\newcommand\Rmj{\ensuremath{\mathbb{R}^{m_j}}}
\renewcommand\l{\ensuremath{\mathcal{L}}}
\renewcommand\le{\ensuremath{\mathcal{L}_e}}
\newcommand\lex{\ensuremath{\mathcal{L}_{ex}}}
\newcommand\ld{\ensuremath{\mathcal{L}_d}}
\renewcommand\lg{\ensuremath{\mathcal{L}_g}}
\newcommand\he{\ensuremath{\mathcal{H}_e}}
\newcommand\hd{\ensuremath{\mathcal{H}_d}}
\newcommand\dt{\ensuremath{\Delta t}}
\newcommand\Me{\ensuremath{\mat{M}_{\le}}}
\newcommand\fe{\ensuremath{\vec{f}_{\le}}}
\newcommand\Pe{\ensuremath{\mat{P}_{\le}}}
\newcommand\M{\ensuremath{\mat{M}}}
\newcommand\Mnh{\ensuremath{\mat{M}_{\textrm{nh}}}}
\newcommand\I{\ensuremath{\mat{I}}}
\newcommand\f{\ensuremath{\vec{f}}}
\newcommand\fnh{\ensuremath{\vec{f}_{\textrm{nh}}}}
\newcommand\h{\ensuremath{\vec{h}}}
\newcommand\Spec{\ensuremath{\mathcal{S}}}
\newcommand\htwo{\ensuremath{\vec{h}_2}}
\newcommand\Md{\ensuremath{\mat{M}_d}}
\newcommand\fd{\ensuremath{\vec{f}_d}}
\newcommand\Mde{\ensuremath{\mat{M}_{de}}}
\newcommand\fde{\ensuremath{\vec{f}_{de}}}
\newcommand\forc{\ensuremath{\vec{\psi}}}
\newcommand\spec{\ensuremath{\left(\M,\f\right)_{\X}}}
\newcommand\x{\ensuremath{\vec{x}}}
\newcommand\xdot{\ensuremath{\dot{\x}}}
\newcommand\xddot{\ensuremath{\ddot{\x}}}
\newcommand\xj{\ensuremath{\vec{x}_j}}
\newcommand\xjdot{\ensuremath{\dot{\xj}}}
\newcommand\xjddot{\ensuremath{\ddot{\xj}}}
\newcommand\xb{\ensuremath{\bar{\vec{x}}}}
\newcommand\xbdot{\ensuremath{\dot{\xb}}}
\newcommand\xbddot{\ensuremath{\ddot{\xb}}}
\newcommand\xt{\ensuremath{\vec{\tilde{x}}}}
\newcommand\xg{\ensuremath{\vec{\x_\textrm{goal}}}}
\newcommand\xtdot{\ensuremath{\dot{\xt}}}
\newcommand\xtddot{\ensuremath{\ddot{\xt}}}
\newcommand\xr{\ensuremath{\x_{\textrm{rel}}}}
\newcommand\xrdot{\ensuremath{\dot{\x}_{\textrm{rel}}}}
\newcommand\xrddot{\ensuremath{\ddot{\x}_{\textrm{rel}}}}
\newcommand\q{\ensuremath{\vec{q}}}
\newcommand\qdot{\ensuremath{\dot{\q}}}
\newcommand\qddot{\ensuremath{\ddot{\q}}}
\newcommand\qt{\ensuremath{\vec{\tilde{q}}}}
\newcommand\qtdot{\ensuremath{\dot{\qt}}}
\newcommand\qtddot{\ensuremath{\ddot{\qt}}}
\newcommand\fk{\ensuremath{\textrm{fk}}}

\newcommand\params{\ensuremath{\vec{\Theta}}}
\newcommand\cost{\ensuremath{c}}
\newcommand\costestimate{\ensuremath{\tilde{\cost}(\params)}}
\newcommand\costdist{\ensuremath{\cost_{\textrm{distance}}}}
\newcommand\costpath{\ensuremath{\cost_{\textrm{path}}}}
\newcommand\costcol{\ensuremath{\cost_{\textrm{clearance}}}}
\newcommand\weight{\ensuremath{w}}
\newcommand\weightdist{\ensuremath{\weight_{\textrm{distance}}}}
\newcommand\weightpath{\ensuremath{\weight_{\textrm{path}}}}
\newcommand\weightcol{\ensuremath{\weight_{\textrm{clearance}}}}
\newcommand\trial{\ensuremath{\mathcal{T}}}
\newcommand\mpp{\ensuremath{\mathcal{MP}}}
\newcommand\paramsmp{\ensuremath{\params_{\mpp}}}
\newcommand\planner{\ensuremath{\mathcal{P}}}
\newcommand\kgeocol{\ensuremath{k_{\textrm{geo,col}}}}
\newcommand\expgeocol{\ensuremath{\beta_{\textrm{geo,col}}}}
\newcommand\kfincol{\ensuremath{k_{\textrm{fin,col}}}}
\newcommand\expfincol{\ensuremath{\beta_{\textrm{fin,col}}}}
\newcommand\kgeolimit{\ensuremath{k_{\textrm{geo,limit}}}}
\newcommand\expgeolimit{\ensuremath{\beta_{\textrm{geo,limit}}}}
\newcommand\kfinlimit{\ensuremath{k_{\textrm{fin,limit}}}}
\newcommand\expfinlimit{\ensuremath{\beta_{\textrm{fin,limit}}}}
\newcommand\kgeoself{\ensuremath{k_{\textrm{geo,self}}}}
\newcommand\expgeoself{\ensuremath{\beta_{\textrm{geo,self}}}}
\newcommand\kfinself{\ensuremath{k_{\textrm{fin,self}}}}
\newcommand\expfinself{\ensuremath{\beta_{\textrm{fin,self}}}}
\newcommand\baseinertia{\ensuremath{m_{\textrm{base}}}}
\newcommand\kattractor{\ensuremath{k_{\textrm{attractor}}}}
\newcommand\Bmax{\ensuremath{\mat{B}_{\textrm{max}}}}
\newcommand\Bmin{\ensuremath{\mat{B}_{\textrm{min}}}}
\newcommand\alphaex{\ensuremath{\alpha_{\textrm{ex}}}}
\newcommand\alphale{\ensuremath{\alpha_{\le}}}
\newcommand\alphab{\ensuremath{\alpha_{\beta}}}
\newcommand\exfactor{\ensuremath{v_{ex}}}
\newcommand\radiusshift{\ensuremath{r_{shift}}}

\newcommand\J{\ensuremath{\mat{J}_{\phi}}}
\newcommand\Jt{\ensuremath{\mat{J}^T_{\phi}}}
\newcommand\Jdot{\ensuremath{\dot{\mat{J}}_{\phi}}}
\newcommand\Jnh{\ensuremath{\mat{J}_{\textrm{nh}}}}
\newcommand\Jnht{\ensuremath{\mat{J}^T_{\textrm{nh}}}}
\newcommand\Jnhdot{\ensuremath{\dot{\mat{J}}_{\textrm{nh}}}}
\newcommand{\map}{\ensuremath{\phi}}
\newcommand{\mapt}{\ensuremath{\phi_t(\vec{q})}}
\newcommand{\mapd}{\ensuremath{\phi_d}}
\newcommand{\g}{\ensuremath{\vec{g}(t)}}
\newcommand{\gp}{\ensuremath{\vec{g}^{\prime}(t)}}
\newcommand{\gpp}{\ensuremath{\vec{g}^{\prime\prime}(t)}}
\newcommand{\pull}[2]{\textrm{pull}_{#1}{#2}}
\newcommand{\energize}[2]{\textrm{energize}_{#1}{#2}}
\newcommand{\der}[2]{\partial_{#1}#2}
\newcommand{\dertwo}[2]{\partial^2_{#1}#2}
\newcommand{\derf}[2]{\frac{\partial#2}{\partial#1}}
\newcommand{\derftwo}[3]{\frac{\partial^2#3}{\partial#1\partial#2}}
\newcommand{\dert}[1]{\frac{d}{dt}#1}
\newcommand{\pinv}[1]{#1^{\dagger}}
\newcommand{\norm}[1]{\left\lVert#1\right\rVert}
\newcommand{\abs}[1]{\lVert#1\rVert}
\newcommand\mat[1]{\ensuremath{\bm{#1}}}
\renewcommand\vec[1]{\ensuremath{\bm{#1}}}
\newcommand\switching[2]{\ensuremath{{\vec{s}_{#1}(#2)}}}
\newcommand\sign[1]{\ensuremath{\textrm{sgn}(#1)}}

\newcommand{\panda}{Panda robot}
\newcommand{\boxer}{Clearpath Boxer}
\newcommand{\ros}{ROS}
\newcommand{\optuna}{Optuna}
\newcommand{\reachingontable}{\textit{reaching-on-table}}
\newcommand{\reachinginring}{\textit{reaching-in-ring}}

\definecolor{OliveGreen}{rgb}{0,0.6,0}
\newcommand{\MS}[1]{\textbf{{\color{OliveGreen}{{#1}}}}} 

\iftrackchanges \newcommand{\changed}[2][]{{\color{blue}{#2}}} \else \newcommand{\changed}[2][]{#2} \fi
\iftrackchanges \newcommand{\deleted}[2][]{} \else \newcommand{\deleted}[2][]{} \fi

\acrodef{mpc}[MPC]{Model Predictive Control}
\acrodef{sf}[SF]{Static Fabrics}
\acrodef{df}[DF]{Dynamic Fabrics}

\title{\LARGE \bf
Autotuning Symbolic Optimization Fabrics for Trajectory Generation
}

\author{Max Spahn* and Javier Alonso-Mora*
\thanks{*This research was supported by Ahold Delhaize. All content represents the opinion of the author(s), which is not necessarily shared or endorsed by their respective employers and/or sponsors.}
\thanks{The authors are with the Department of Cognitive Robotics, Delft University of Technology, 2628 CD, Delft, The Netherlands
       {\tt\small \{m.spahn, j.alonsomora\}@tudelft.nl}}
}

\maketitle
\thispagestyle{empty}
\pagestyle{empty}


\begin{abstract}
In this paper, we present an automated parameter optimization method for
trajectory generation. We formulate parameter optimization as a constrained
optimization problem that can be effectively solved using Bayesian
optimization. While the approach is generic to any trajectory generation
method, we showcase it using optimization fabrics. Optimization fabrics are a
geometric trajectory generation method based on non-Riemannian geometry. By
symbolically pre-solving the structure of the tree of fabrics, we obtain a
parameterized trajectory generator, called symbolic fabrics. We show that
autotuned symbolic fabrics reach expert-level performance in a few trials.
Additionally, we show that tuning transfers across different robots, motion
planning problems and between simulation and real world. Finally, we
qualitatively showcase that the framework could be used for coupled mobile
manipulation.
\end{abstract}

%
%
\vspace{-1em}
Code \href{https://github.com/maxspahn/optuna_fabrics}{github.com/maxspahn/optuna\textunderscore fabrics}

\section{Introduction}
\label{sec:intro}
Mobile manipulation is the field of robotics concerned with highly capable robots 
characterized by their locomotion and manipulation ability. Such robots
are getting ever more attention as they will be deployed to human-shared environments, 
like households or warehouses. In such dynamic environments, fast trajectory generation
is crucial to avoid collisions and react quickly to changing goal definitions. 

Trajectory generation is often addressed by solving an optimization problem
that consists of a scalar objective function -- the dynamics or transition
function -- and several constraints. As the degrees of freedom and number of
constraints increase, solving that problem in real-time becomes challenging.
This is especially limiting in the case of mobile manipulation
\cite{spahn2021coupled}. Optimization fabrics represent a different approach to
the problem, as they formulate \changed{trajectory generation} as the
shortest-geodesic-problem in a manifold of the configuration space
\cite{Karaman2011}. 

With optimization fabrics, different components, or desired behaviors, such as
collision avoidance and joint limit avoidance, are combined using Riemannian
metrics. As the structure of the resulting trajectory generation methods
remains unchanged across all time steps, it can be composed before runtime,
thus saving computational costs during executing. 
Optimization fabrics, but also their predecessor Riemannian Motion Policies
(RMPs), have shown impressive results for several manipulator applications,
including dynamic and crowded environments \cite{Wyk2022,Xie2021,Spahn2022}.
However, despite their theoretical properties of inherent collision avoidance
and convergence, these methods require expertise and intuition to tune
individual components to generate smooth and well-behaving trajectories. 
\begin{figure}
    \centering
    \includegraphics[width=0.8\linewidth]{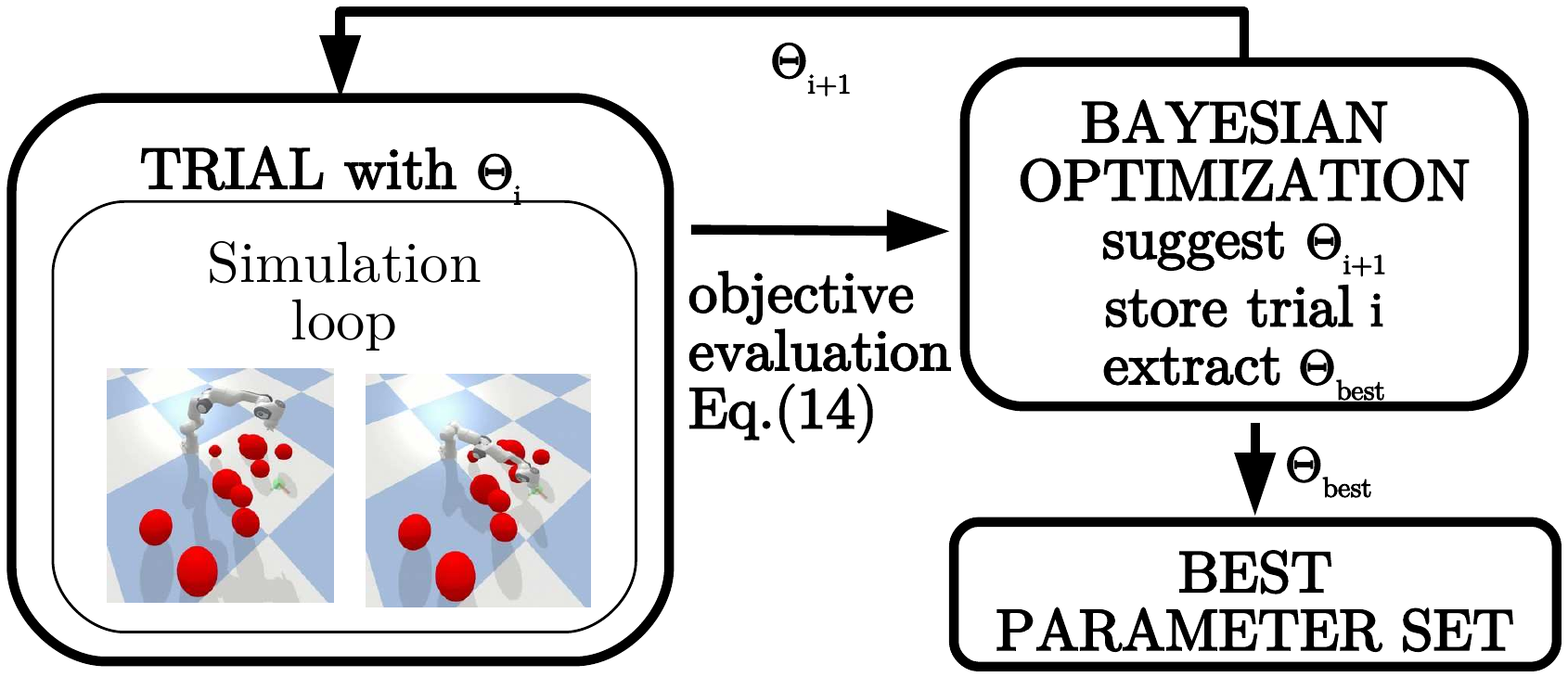}
    \captionsetup{belowskip=-20pt}
    \caption{
      Overview of one trial in the tuning pipeline for symbolic optimization
      fabrics. The objective function is evaluated after an entire trial run is
      simulated. Using Bayesian optimization, a new parameter set is suggested
      based on the history of trials. The best parameter set is extracted from
      all trials.
    }
    \label{fig:overview}
\end{figure}

\textit{Contributions:}
To address this issue, we formulate optimization fabrics as a \textbf{symbolic
trajectory generation} method. Precisely, the combination of the individual
components (joint limit avoidance, goal reaching, collision avoidance, etc.) is
performed in a parameterized way before runtime. Separating composition and
evaluation allows for changing the individual parameters at runtime while
achieving low computational costs. Additionally, this allows formulating
parameter-tuning as a constrained optimization problem. Solving this problem
effectively \textbf{automates the tuning process} systematically using
Bayesian optimization. We show that automated tuning requires only few trials
to achieve similar performance to an expert in the field, and systematically
outperforms a randomized parameter setting. Moreover, we show that one
parameter tuning generalizes across different robots, to some extent, across
different tasks and between simulation and real world.
Finally, we demonstrate how \textbf{coupled mobile manipulation} with
a differential drive can be achieved using autotuned optimization fabrics
for in-store order-picking integrating visual servoing.

\section{Related Works}%
\label{sec:related_works}
\subsection{Geometric control for trajectory generation}
Operational space control was the first control method that imposed
a desired dynamical system onto a robotic system \cite{Khatib1985,Khatib1987}.
The concept was an important step toward naturally controlling 
kinematically redundant robots. The concept was formalized in the field 
of geometric control, where the study of differential geometry leads to 
stable and converging behavior under geometric conditions \cite{bullo2019geometric}.
More recently, RMPs for manipulation tasks offered
a highly reactive trajectory generation method \cite{Ratliff2018,Cheng2019}. This method
achieves composable behavior by introducing a split between the importance metric
and the forcing term. Using the \textit{pullback} and \textit{pushforward} operator
to change between manifolds of the configuration space, individual 
components, such as collision avoidance and goal attraction, can be designed
iteratively. However, RMPs require intuition and 
experience when being designed, and convergence 
can only be proven conditionally \cite{Ratliff2020}. Later, optimization fabrics
were introduced that are able to completely decouple importance metrics
and the defining geometry. Under simple construction rules for these 
two components, convergence can be easily guaranteed \cite{Ratliff2020,Xie2021,Li2021,meng2019neural}.
In our prior work on optimization fabrics, the framework was first applied to
mobile manipulation and generalized to more dynamic environments. \cite{Spahn2022}.
\begin{figure}[t]
    \centering
    \includegraphics[width=0.8\linewidth]{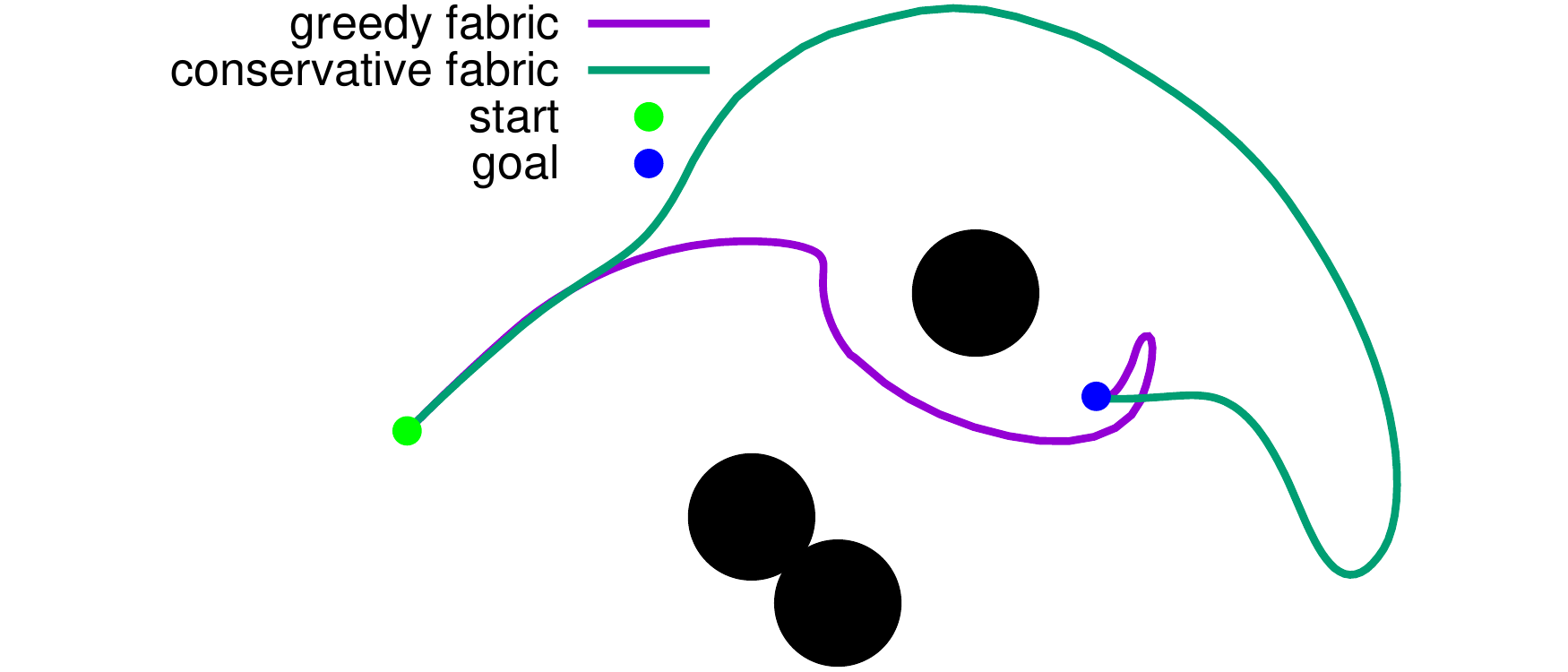}
    \captionsetup{belowskip=-10pt}
    \caption{Two different parameter sets for optimization fabrics given the same problem. While the greedy tuning
    is more aggressive (purple), the more conservative tuning results in a smoother trajectory (green).}
    \label{fig:effect_tuning}
    \vspace{-10pt}
\end{figure}

\subsection{Autotuning for trajectory generation}
Autotuning can be beneficial for trajectory generation when using model predictive control.
In \cite{loquercio_autotune_2022}, an autotuned
model predictive controller has outperformed a manual tuned controller of the same kind by 
25\%. Jointly optimizing parameters and the model of the controller, \textit{AutoMPC} 
showed the benefit of parameter tuning in the context of simultaneous system
identification and control \cite{edwards_automatic_2021}. These methods are explicitly 
formulated for model predictive control and do not transfer easily to other trajectory generation
methods. In contrast, we propose a generic parameter optimization approach to trajectory generation.

\subsection{Hyperparameter tuning in machine learning}
Within the machine-learning field, hyperparameter tuning has shown to be highly 
important for all different kinds of applications \cite{yang2020hyperparameter,hutter_automated_2019,optuna}. 
Parameter optimization aims to minimize 
training costs while achieving the best possible performance. Hyperparameter tuning 
is most valuable in extremely
costly applications such as reinforcement learning \cite{zoph_neural_2017}. Generally, two different search
algorithms have been investigated: grid search and random search \cite{bergstra_random_nodate}.
Current state-of-the-art methods for parameter search are based on 
random search with a Bayesian optimizer \cite{optuna,bergstra_algorithms_nodate}.
While the machine-learning community has largely agreed on the importance of
parameter tuning, systematic tuning of trajectory generation methods are not well established. 
In this paper, we showcase, with the example of optimization fabrics, how important 
parameter tuning is and how trajectory generation can benefit from it.

\section{Overview}
\label{sec:overview}
In this paper, we first recall very briefly the theory of optimization fabrics and the steps
to use it for trajectory generation (\cref{sec:optimization_fabrics}). Then, we formulate 
optimization fabrics as a symbolic trajectory generator, so that combining
of individual components is only performed once (\cref{sec:symbolic_fabrics}).
Then, we formulate parameter tuning for trajectory generation as a constrained optimization problem
and propose Bayesian optimization for effective autotuning (\cref{sec:tuning}).  
As an example, we apply this autotuning to symbolic optimization
fabrics (\cref{sec:experimental_results}), but it is generally 
independent of the trajectory generator at hand.

\section{Background}
\label{sec:optimization_fabrics}
In this section, we very briefly introduce the concepts required for trajectory
generation with optimization fabrics. For a more in-depth introduction to
optimization fabrics and its foundations in differential geometry, the reader
is referred to \cite{Ratliff2020,Spahn2022,Wyk2022}.
\subsection{Configurations and task variables}%
\label{sub:configurations_and_task_variables}
We denote $\q\in\Q\subset\Rn$ a
configuration of the robot with $n$ its degrees of freedom;
\Q{} is the configuration space of the generalized coordinates
of the system. Generally, $\q(t)$ defines the robot's configuration at time $t$, so that 
\qdot, \qddot{} define the instantaneous derivatives of the robot's configuration.
Similarly, we assume
that there is a set of task variables $\xj\in\Xj\subset\Rmj$ with variable dimension
$m_j \leq n$. The task space \Xj{} defines an arbitrary manifold of the configuration
space \Q{} in which a robotic task can be represented. 
Further, we assume that there is a differential map
$\map_j:\Rn\rightarrow\Rmj$ that relates the configuration space to the $j^{th}$ task
space. For example, when a task variable is defined as the end-effector position, then
$\map_j$ is the positional part of the forward kinematics. On the other hand, if a task
variable is defined to be the joint position, then $\map_j$ is the identity function. 
In the following, we drop the subscript $j$ in most cases for readability when the
context is clear.

We assume that \map{} is in $\mathcal{C}^1$ so that the Jacobian is
defined as
\begin{equation}
  \J = \derf{\q}{\map} \in \mathcal{R}^{m\times n}, 
\end{equation}
or $\J = \der{\q}{\map}$ for short.
Thus, we can write the total time derivatives of \x{} as
$\xdot = \J\qdot$ and $\xddot = \J\qddot + \Jdot\qdot$.
\subsection{Spectral semi-sprays}%
\label{sub:semi_spectral_sprays}
Inspired by simple mechanics (e.g., the simple pendulum), the framework of optimization
fabrics designs motion policies as second-order dynamical
systems $\xddot = \pi(\x,\xdot)$~\cite{Cheng2020,Ratliff2020}.
The motion policy is defined by the differential equation
$\M\xddot + \f = 0$, where $\M(\x,\xdot)$ and $\f(\x,\xdot)$ are functions of position and
velocity. Besides, \M{} is symmetric and invertible. We denote such systems as $\Spec = \spec$ and
refer to them as \textit{spectral semi-sprays}, or \textit{specs} for short.  When the space of
the task variable is clear from the context, we drop the subscript. 

\subsection{Operations on specs}%
\label{sub:operations_on_specs}
Complex trajectory generation is composed of multiple components, such as collision avoidance, joint limits
avoidance, etc. The power of optimization fabrics lies in the metric-weighted sum to
combine multiple components from different manifolds.
These operations are derived from operations on specs and are briefly recalled here.

Given a differential map $\map: \Q\rightarrow\X$ and a spec \spec{}, the \textit{pullback}
is defined as 
\begin{equation}
  \pull{\map}{\spec} = {\left(\Jt\M\J, \Jt(\f+\Jdot\qdot)\right)}_{Q}.
\end{equation}
The pullback allows converting between two distinct manifolds (e.g. a spec could be 
defined in the robot's workspace and pulled into the robot's configuration space using
the pullback with \map{} being the forward kinematics).

For two specs, $\Spec_1 = {\left(\M_1,\f_1\right)}_{\X}$ and 
$\Spec_2 = {\left(\M_2,\f_2\right)}_{\X}$, their \textit{summation} is defined by:
\begin{equation}
  \Spec_1 + \Spec_2 = {\left(\M_1 + \M_2, \f_1 + \f_2\right)}_{\X}.
  \label{eq:specs_summation}
\end{equation}

Additionally, a spec can be \textit{energized} by a Lagrangian energy. Effectively, 
this equips the spec with a metric.
Specifically, given a spec of form $\Spec_{\vec{h}} = (\mat{I},\vec{h})$ and 
an energy Lagrangian \le{} with the derived equations of motion $\M_{\le}\xddot + \f_{\le} =0$, 
we can define the operation
\begin{equation}
  \begin{split}
  S_{\vec{h}}^{\le} &= \text{energize}_{\le}\{S_{\vec{h}}\} \\
    &= (\Me, \fe + \Pe[\Me\vec{h} - \fe]), 
  \end{split}
  \label{eq:energization}
\end{equation}
where $\Pe = \Me\left(\Me^{-1} - \frac{\xdot\xdot^T}{\xdot^T\Me\xdot}\right)$ is an
orthogonal projector. The resulting spec is an \textit{energized spec} and 
we call the operation \textit{energization}.

With spectral semi-sprays and the presented operations,
avoidance behavior, such as joint limit avoidance, collision
avoidance or self-collision avoidance, can be realized.

\subsection{Optimization fabrics}%
\label{sub:optimization_fabrics}
In the previous subsection, we explained how different avoidance behaviors can
be combined. Spectral semi-sprays can additionally be \textit{forced} by a
potential, denoted as the \textit{forced variant} of form $\Spec_{\forc} =
\left(\M,\f + \der{\x}{\forc}\right)$. This forcing term clearly changes the
behavior of the system. Optimization fabrics introduce construction rules to
make sure that the solution path $\x(t)$ of $\Spec_{\forc}$ converges towards
the minimum of $\forc(\x)$. Then, the potential is designed in such a way that
its minimum represents a goal state of the motion planning problem.

First, the initial spec that represents an avoidance component
is written in the form $\xddot + \vec{h}(\x,\xdot) = 0$,
such that $\vec{h}$ is \textit{homogeneous of degree 2}:
$\vec{h}(\x,\alpha\xdot) = \alpha^2\vec{h}(\x,\xdot)$ 
(\textbf{Creation}). 
Secondly, the geometry is
energized (\cref{eq:energization})
with a Finsler structure~\cite[Definition 5.4]{Ratliff2020} (\textbf{Energization}).
The property of homogeneity of degree 2 and the energization with the Finsler structure
guarantees, according to ~\cite[Theorem 4.29]{Ratliff2020}, that the energized spec
forms a \textit{frictionless fabric}.
A frictionless fabric is defined to optimize the forcing potential \forc{} when being
damped by a positive definite damping term~\cite[Definition 4.4]{Ratliff2020}.
Thirdly, all avoidance components are combined in the configuration space
of the robot using the pullback and summation operation (\textbf{Combination}).
Note, that both operations are closed 
under the algebra designed by these operations, i.e. every pulled optimization fabric or the sum
of two optimization fabrics is, itself, an optimization fabric.
In the last step, the combined spec is forced by the potential \forc{} with the desired minimum and
damped with a positive definite damping term (\textbf{Forcing}).
This resulting system of form $\M\qddot + \f(\q, \qdot) + \der{\q}{\forc} + \beta\qdot = 0$ is solved to 
obtain the trajectory generation policy in acceleration form $\qddot = \pi(\q,\qdot)$.

\section{Symbolic fabrics}
\label{sec:symbolic_fabrics}
A trajectory generator that is based on optimization fabrics is composed of several components, such as collision
avoidance, joint limit avoidance, goal attraction, etc. Each component contributes
to the resulting optimization fabric through the metric-weighted summation that creates the tree of fabrics.
The trajectory generator is parameterized by the individual terms of the components.
Here, we lay out the parameterization for collision
avoidance, joint limit avoidance, self-collision avoidance, and speed-control.
In our framework, the tree of
fabrics is generated before runtime as a symbolic expression, to which the parameters are set at runtime.
Note that the approach of symbolic pre-solving results in much higher planning frequencies.
In the following, we explain the individual parameters that we exposed symbolically. The form of the 
individual terms is adapted from \cite{Ratliff2020,Xie2021,Wyk2022} but written in a symbolic form.

\paragraph{Basic inertia}
The final tree of fabrics is equipped with a basic inertia metric that indicates how reactive the entire 
motion is. This basic inertia metric is derived from the symbolic Finsler structure:
$\le = 0.5\baseinertia\qdot^{T}\I\qdot$.

\paragraph{Collision avoidance}
For collision avoidance, the task manifold \X{} is defined by the distance function between an obstacle and 
a robot link. The differential map used is defined as
\[
    \map_{\textrm{i}}(\q) = \frac{\norm{\fk_{\textrm{i}}(\q) - \x_{\textrm{obst}}}}{r_{\textrm{obst}} + r_{\textrm{i}}} - 1,
\]
where $\fk_{\textrm{i}}(\q)$ is the positional forward kinematics for link $i$ in a configuration \q{},
$r_{\textrm{obst}}$ and $r_{\textrm{i}}$ are the radii of the englobing
spheres for the obstacle and the link respectively. 
While this mapping between configuration space and task manifold is different for each obstacle and each
collision link of the robot, the geometry and metric are the same for all of them.
For the geometry 
$\xddot + \vec{h}(\x,\xdot) = \vec{0}$, we use the parameterized forcing term
\begin{equation}
    \vec{h}(\x,\xdot) = \frac{-\kgeocol}{\x^{\expgeocol}}\xdot^2,
\end{equation}
where \kgeocol{} and \expgeocol{} are parameters of the trajectory generator. Generally, we use $k$ and $\beta$ for
proportional parameters and exponential parameters.
The Finsler structure for collision avoidance is parameterized as 
\begin{equation}
    \le(\x,\xdot) = \frac{\kfincol}{\x^{\expfincol}} \left(-0.5 (\sign{\xdot} - 1)\right) \xdot^2,
\end{equation}
where $\sign{\xdot}$ is the signum-operator returning the sign of \xdot{}.

\paragraph{Self-collision avoidance}
For self-collision avoidance, the task manifold \X{} is 
defined similarly to collision avoidance:
\[
    \map_{i,j} = \frac{\norm{\fk_i(\q) - \fk_j(\q)}}{r_i + r_j} - 1,
\]
where $\fk_i(\q)$ and $\fk_j(\q)$ are the positional forward kinematics of the two links
for a self-collision pair and $r_i$ and $r_j$ are the 
radii for both englobing spheres.
The geometries are defined analogously 
\begin{equation}
    \vec{h}(\x,\xdot) = \frac{-\kgeoself}{\x^{\expgeoself}}\xdot^2.
\end{equation}
The Finsler structure for collision avoidance is parameterized as 
\begin{equation}
    \le(\x,\xdot) = \frac{\kfinself}{\x^{\expfinself}} \left(-0.5 (\sign{\xdot} - 1)\right) \xdot^2.
\end{equation}

\paragraph{Joint limit avoidance}
For joint-limit avoidance, two simple differential maps
denoting the distance to the joint limits are used, specifically
\begin{align*}
    \map_{\textrm{limit,i,lower}}(\q) &= \q_{i} - \q_{\textrm{min,i}},\forall i \in (1,\dots,n)\\
    \map_{\textrm{limit,i,upper}}(\q) &= \q_{\textrm{max,i}} - \q_i,\forall i \in (1,\dots,n).
\end{align*}
Similar to collision avoidance, we use the parameterized forcing term
\begin{equation}
    \vec{h}(\x,\xdot) = \frac{-\kgeolimit}{\x^{\expgeolimit}}\xdot^2
\end{equation}
and the Finsler structure
\begin{equation}
    \le(\x,\xdot) = \frac{\kfinlimit}{\x^{\expfinlimit}} \left(-0.5 (\sign{\xdot} - 1)\right) \xdot^2.
\end{equation}

%
\paragraph{Speed control}
As the root of the tree of fabrics is a frictionless fabric, it only converges if damped \cite{Ratliff2020}. 
Constant damping is sufficient to achieve the theoretical properties that are needed for trajectory generation.
However, \cite{Ratliff2020,Wyk2022,Ratliff2021} proposed enhanced damping under the name of \textit{speedcontrol}. 
We employ the same damping strategy while adding parameterization. The technique is based on a
dynamic damping modification based on the distance to the goal. Specifically, the final optimization fabric
is damped according to 
\[
    \qddot = -\vec{h}_2 - \M^{-1}\der{\q}{\forc} + \alphaex\qdot - \beta\qdot, 
\]
where $\vec{h}_2$ is the sum of all pulled forcing terms, \M{} is the sum of 
all metrics of the individual geometries,
$\der{\q}{\forc}$ is the goal attraction term pulled in the configuration space, 
$\alphaex$ is a weighted sum of $\alphaex^0$
that maintains constant execution energy without goal attraction
and $\alphaex^{\forc}$ that maintains constant execution
energy with goal attraction:
\[
    \alphaex = \switching{\eta}{\lex}\alphaex^0 + (1-\switching{\eta}{\lex})\alphaex^{\forc}.
\]
Then, $\beta$ is the damping term, computed as:
\[
    \beta = \switching{\beta}{\q}\Bmax + \Bmin + \max(0, \alphaex - \alphale),
\]
where \Bmax{} and \Bmin{} are the upper and lower damping values and
\alphale{} is the energization coefficient maintaining
constant system energy (not execution energy) without goal attraction. 
The switching functions \switching{\beta}{\q}, \switching{\eta}{\q} are further parameterized as
\begin{align*}
    \switching{\beta}{\q} &= 0.5(\tanh{-\alphab(\abs{\q} - \radiusshift)) + 1}\\
    \switching{\eta}{\lex} &= 0.5(\tanh{\left(-0.5 \lex (1 - \exfactor) - 0.5\right)} + 1),
\end{align*}
where \radiusshift{} determines the distance to the goal at which the switch between \Bmin{} and \Bmax{} occurs,
\alphab{} is the steepness of that switching, \lex{} is the user-defined execution energy (usually a simple
kinetic energy in joint space) and \exfactor{} is the execution energy factor, i.e. it determines the desired speed
of motion.
For a detailed discussion on speed control with optimization fabrics, we refer
to previous works on optimization fabrics \cite{Wyk2022,Ratliff2020}.

We group all parameters resulting from the symbolic fabrics defined here into 
a vector of parameters \params{}. 
All parameters are listed in \cref{tab:search_space}.

\section{Parameter tuning as an optimization problem}
\label{sec:tuning}
We define parameter tuning as a constrained optimization problem:
\begin{equation}
    \params^{\ast} = \arg\min_{\params}\cost(\params),\ 
    \textrm{s.t}\ \params_{\textrm{min}} < \params < \params_{\textrm{max}},
    \label{eq:optimization_problem}
\end{equation}
where $\params_{\textrm{max}}$ and $\params_{\textrm{min}}$ are the upper and lower bounds of the parameters.
The objective $\cost(\params)$ is a function of the parameters specifying the tree of fabrics and can 
be evaluated after one trajectory planning problem has finished. We call the evaluation of one parameter set
a \textit{trial}.
Next, we propose an objective function that is flexible
as different scenarios may require different parameter tuning.
\subsection{Objective}
The objective function $\cost(\params)$ is a weighted sum of
several metrics, that are invariant to the robot:
\begin{equation}
    \cost(\params) = 
      \weightdist\costdist
    + \weightpath\costpath
    + \weightcol\costcol.
    \label{eq:optimization_objective}
\end{equation}

\costdist{} accounts for the normalized, summed distance to the goal over one trial and is defined as
\begin{equation}
    \costdist = \frac{\sum_{i=0}^{T} \norm{\x_i - \xg}}{\norm{\x_0 - \xg}},
\end{equation}
where $i\in[0, T]$ are the discretized time steps and $\xg$ is the goal of the motion planning problem.
\costpath{} accounts for the normalized path length over one trial and is defined as
\begin{equation}
    \costpath = \frac{\sum_{i=1}^{T} \norm{\x_i - \x_{i-1}}}{\norm{\x_0 - \xg}}.
\end{equation}
\costcol{} accounts for the average clearance to obstacles over one trial and is defined as
\begin{equation}
    \costcol = \frac{1}{T}\sum_{i=1}^{T} \min_{o^j}\norm{\x_i - o^j_i)}, 
\end{equation}
where $o^j_i$ is the position of obstacle $j$ at time step $i$.
Each of these terms is evaluated after an entire trial that was obtained by a specific set of parameters.
\begin{algorithm}
\SetAlgoLined
Formulate trajectory generator with parameters $\params$\\
Define parameter space by $\params_{\textrm{min}},\params_{\textrm{max}}$\\
Formulate objective $\cost(\params)$\\
Initialize objective function estimate \costestimate{} \\
\For{i = 0 to N}{
    Suggest parameter $\params_i$ based on \costestimate{}\\
    \For{t = 0 to T}{
        Compute action with parameter set $\params_i$\\
        Apply action to robot\\
        Store observation relevant for metrics\\
    }
    Evaluate $\cost(\params_i)$\\
    Update \costestimate{}\\
}
Extract the best parameter set $\params_{\textrm{best}}$\\
\caption{Autotuning for trajectory generators}
\label{algo:autotuning}
\end{algorithm}
\subsection{Bayesian optimization}
\label{sub:bayesian_optimization}
In the tuning phase, the problem specification for the investigated scenario,
e.g., the goal and obstacle positions, across all trials during tuning remains
the same while \params{} are optimized according to the objective. To solve the
Bayesian optimization we employ the \textit{Tree-structured Parzen Estimator} as
it has shown improved performance over grid-search and conventional random
search in machine learning applications
\cite{turner2021bayesian,bergstra_algorithms_nodate}. To deploy this technique
we used \optuna{}, a hyperparameter optimization framework initially designed
for machine learning applications \cite{optuna}. The general setup for one trial
is shown in \cref{fig:overview} and the procedure is summarized in
\cref{algo:autotuning}. 
\section{Experimental Results}
\label{sec:experimental_results}
We showcase our parameter optimization method for symbolic fabrics. The search space 
for the parameters is summarized in \cref{tab:search_space}.
\begin{table}[]
    \centering
    \begin{tabular}{c|c|c|c|c}
        Parameter & boundaries & type & distribution & manual \\
        \hline
        \baseinertia{} & $[0,1]$ & float & uniform & 0.2\\ 
        \kgeocol{} & $[0.01,1]$ & float & log & 0.03\\ 
        \kgeolimit{} & $[0.01,1]$ & float & log & 0.3\\ 
        \kgeoself{} & $[0.01,1]$ & float & log & 0.03\\ 
        \kfincol{} & $[0.01,1]$ & float & log & 0.03\\ 
        \kfinlimit{} & $[0.01,1]$ & float & log & 0.05\\ 
        \kfinself{} & $[0.01,1]$ & float & log & 0.03\\ 
        \expgeocol{} & $[1,5]$ & int & uniform & 3\\ 
        \expgeolimit{} & $[1,5]$ & int & uniform & 2\\ 
        \expgeoself{} & $[1,5]$ & int & uniform & 3\\ 
        \expfincol{} & $[1,5]$ & int & uniform & 3\\ 
        \expfinlimit{} & $[1,5]$ & int & uniform & 3\\ 
        \expfinself{} & $[1,5]$ & int & uniform & 3\\ 
        \alphab{} & $[0, 1]$ & float & uniform & 0.5\\
        \Bmin{} & $[0,1]$ & float & uniform & 0.01 \\
        \Bmax{} & $[5, 20]$ & float & uniform & 6.5 \\
        \radiusshift{} & [0.01, 0.1] & float & uniform & 0.05 \\
        \exfactor{} & [1.0, 30] & float & uniform & 15.0 \\
    \end{tabular}
    \
    \caption{Search space for parameters. Some parameters are restricted to integers, and for some
    a log-distribution is applied.}
    \label{tab:search_space}
    \vspace{-15pt}
\end{table}
We first analyze
the importance of tuning for optimization fabrics on the performance of trajectory
generation.
Then, we investigate
how tuned parameters can be transferred across
different robots (\cref{sub:cross_validation_robots}),
different scenarios (\cref{sub:cros_validation_tasks}),
and between simulation and real world (\cref{sub:cross_validation_transfer_real_world}).

\subsection{Experimental setup}
\begin{figure}
    \centering
    \begin{subfigure}[b]{0.35\linewidth}
        \centering
        \includegraphics[width=0.8\textwidth]{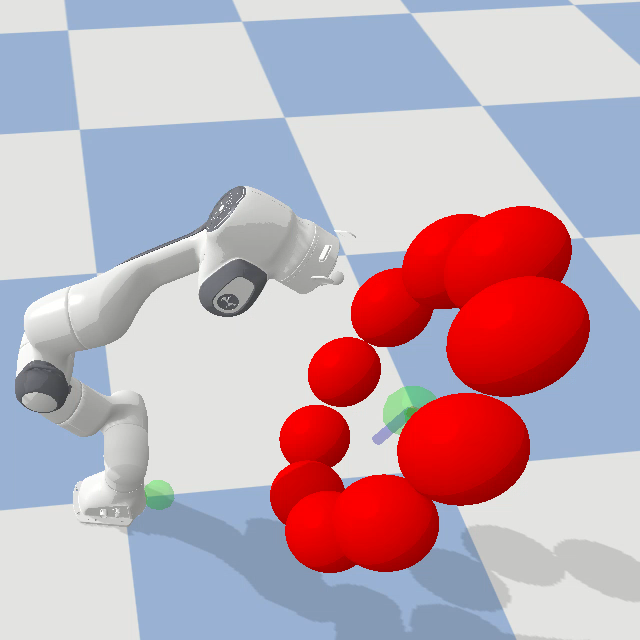}
        \caption{\reachinginring{}}
        \label{fig:reachinginring}
    \end{subfigure}
    \begin{subfigure}[b]{0.35\linewidth}
        \centering
        \includegraphics[width=0.8\textwidth]{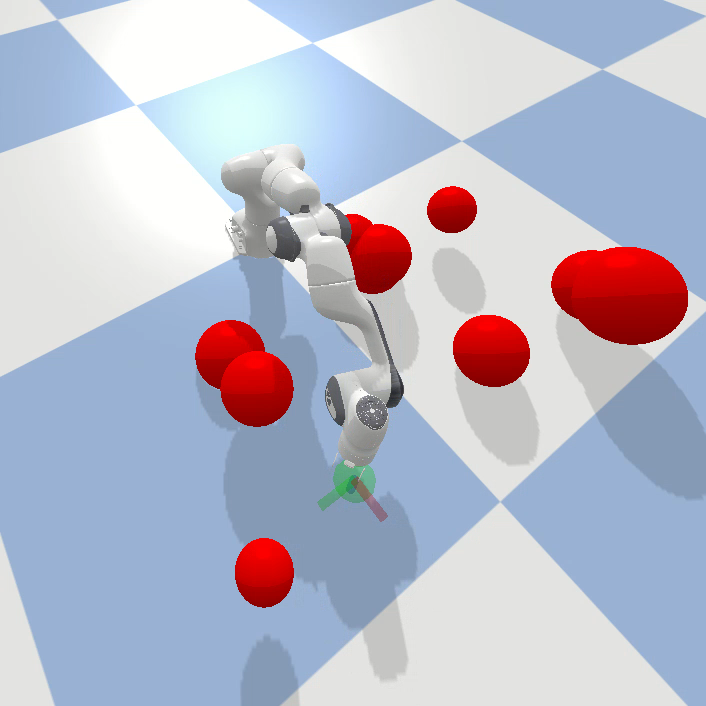}
        \caption{\reachingontable{}}
        \label{fig:reachingontable}
    \end{subfigure}
    \label{fig:tasks}
\end{figure}

\begin{figure}
    \centering
    \begin{subfigure}[b]{0.49\linewidth}
        \includegraphics[width=1\textwidth]{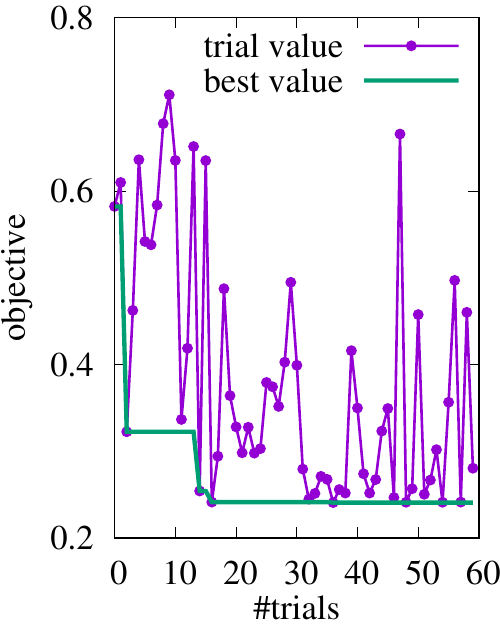}
    \end{subfigure}
    \begin{subfigure}[b]{0.49\linewidth}
        \includegraphics[width=1\textwidth]{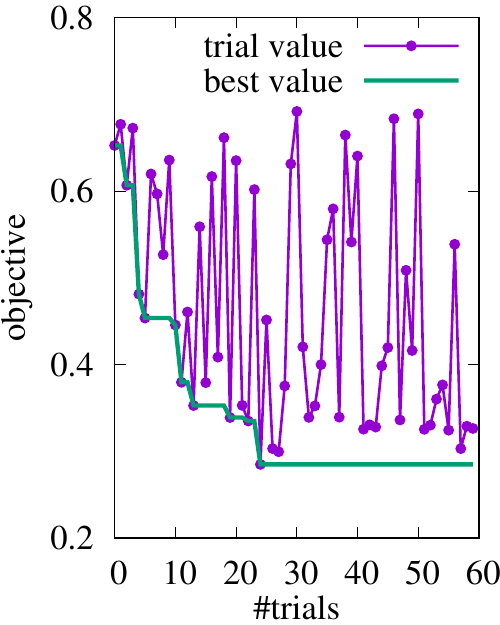}
    \end{subfigure}
    \captionsetup{belowskip=-20pt}
    \caption{Optimization history for simulation (left) and real world (right) for panda robot in 
        \reachinginring{} scenario.}
    \label{fig:history}
\end{figure}

The method was tested in simulation and in the real world on a \panda{} and a
mobile manipulator composed of a \boxer{} and a \panda{}. The simulation uses
the pybullet physics engine with an interface through OpenAI-gym
\cite{spahn_urdf_environment}. 
\deleted{
  The interface to the real world is implemented
  for \ros{} and will be made publicly available alongside this paper.
}
The
different motion planning goals evaluated in this paper are: (a) reaching an
end-effector pose inside a ring of obstacles (\cref{fig:reachinginring})
(similar to the experiment in \cite{Wyk2022}) and (b) reaching an end-effector
pose above a surface with random obstacles (\cref{fig:reachingontable}). The
two scenarios will be referred to as \reachinginring{} and \reachingontable{},
see \cref{fig:tasks}. Unless stated otherwise, the weights are set to
$\weightpath=0.1, \weightcol=0.2, \weightdist=0.7$. We also use this weighted
sum as the performance metric. 
\changed[reviewer6]{
  While these weights are chosen arbitrarily in this work to demonstrate
  the usefulness of autotuning, they should be derived from a human evaluator
  in a more realistic scenario.
}
We refer with \textit{manual} to an expert-tuning, see \cref{tab:search_space}
for specific parameters. During testing, the trial was randomized with changing
obstacles and goals. For autotuning on the robotic arms, we consistently used
$N=60$ trials, although the best parameter set is usually reached earlier, see
\cref{fig:history}.

\subsection{Importance of tuning}
\label{sub:importance_tuning}
\begin{figure}
    \centering
    \includegraphics[width=\linewidth]{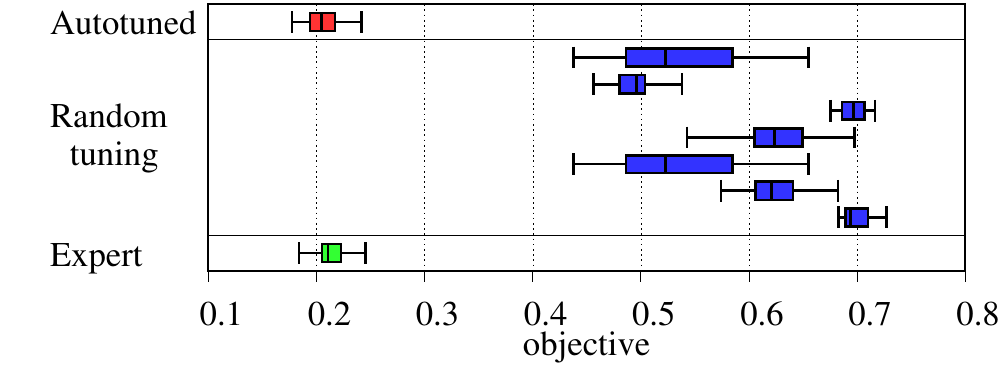}
    \caption{Evaluation for scenario \reachinginring{} autotuned parameters and compared
        to random parameter selection and manual tuning. Autotuning is able to 
        systematically outperform random parameter sets and reach expert level tuning.}
    \label{fig:results_manual_random_autotune}
\end{figure}

We compare the autotuned parameters with seven random parameter sets from the
search space and a manually tuned parameter set that we obtained through
expertise in previous works like \cite{Spahn2022}.
In this experiment, tuning and testing are performed on the test scenario
\reachinginring{}. \deleted{It can be seen that }Tuning is crucial for optimization
fabrics, as the performance with a random parameter set cannot compete with
tuning, \cref{fig:results_manual_random_autotune}. This result was expected and
should only demonstrate that the right parameter set is required to deploy this
method. Autotuned parameters reach a similar performance to the expert. This
result highlights the importance of tuning for optimization fabrics and shows
that autotuning is an effective way to obtain parameter sets for novice users
of optimization fabrics.

\begin{figure}
    \centering
    \includegraphics[width=\linewidth]{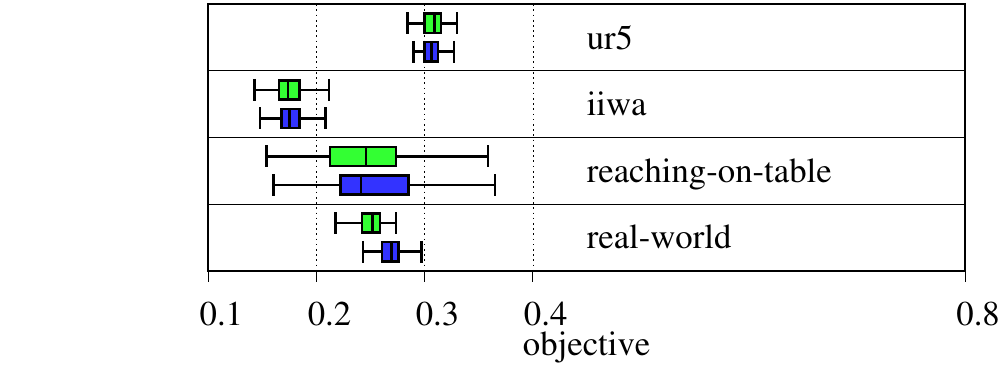}
    \captionsetup{belowskip=-20pt}
    \caption{
        The autotuned for the panda robot in simulation for the
        \reachinginring{} scenario on modified scenarios (blue) is compared to
        autotuned parameter sets obtained on these scenarios directly (green).
        Exchanging the robot (ur5, iiwa) and changing the scenario
        (\reachingontable{}) results in a very small loss in performance, while
        the loss is higher when parameters are transferred between simulation
        and real world (real-world).
    }
    \label{fig:cross_evaluation}
\end{figure}
\subsection{Cross validation: Transfer across robots}
\label{sub:cross_validation_robots}
Without any retuning, we deploy the symbolic optimization fabrics planner tuned
on the \panda{} on two other robots with similar specifications (Kuka LBR IIwa
7, Universal Robot UR5) and compare the performance with tuning performed on
the respective robot. 
\changed[reviewer6]{
  Specifically, we do not change the leaf geometries and energies but change
  differential maps according to relevant collision links on the robot at hand.
}
From \cref{fig:cross_evaluation}, we conclude that tuning is independent of the
robot. This can be explained by the fact, that optimization fabrics are a
purely geometric approach to \changed{trajectory generation} and the different
dimension of the robots do not change the dynamical system enforced onto the
robot.
\subsection{Cross Validation: Transfer across scenarios}
\label{sub:cros_validation_tasks}
In the third experiment, we evaluate how well an autotuned parameter set
transfers to a different scenario. In the specific example, we use the tuning
obtained from the \reachinginring{} case and test it on \reachingontable{}.
Performance can be transferred smoothly if the objective remains the same, see
\cref{fig:cross_evaluation}. However, note that different scenario might
require generally slower motion because of a more crowded environment. Such a
step would require to retune the parameters according to the new objective.
\subsection{Cross Validation: Transfer real world}
\label{sub:cross_validation_transfer_real_world}
As optimization fabrics are a geometric method \cite{Wyk2022}, they should be
independent of the robot embodiment. Relying on the low-level controller. In
this paper, we investigate how the performance is affected by the transfer from
the simulation environment to the real world. 
\deleted{
  For the real-world experiments,
  we make use of the \textit{fabrics-ros-bridge} published alongside this work.
}
Performance benefits from tuning in the real world highlight that low-level
controller differences affect the behavior, see \cref{fig:cross_evaluation}.
Specifically, the accumulated distance to the goal is increased ($0.14$m tuned
in the real world vs $0.16$m tuned in simulation) when tuning is transferred
between simulation and real world. Thus, there is added value in tuning in the
real world. Our framework offers to quickly tune fabrics in the real-world
using the \textit{fabrics-ros-bridge}. With relaxed performance requirements, it
is sufficient to tune in simulation.%

\subsection{Cross Validation: Transfer mobile manipulator}
Finally, we qualitatively test the performance of the tuning method on a real
mobile manipulator with 10 degrees of freedom. After only $N=30$ trials, the
robot was able to perform coupled mobile manipulation based on a visual
servoing approach \cite{chaumette2016visual}. Symbolic optimization fabrics are
especially suited for visual servoing as their symbolic character allows them
to constantly update the position of the goal. A video of this experiment is
attached to the paper.

\begin{figure}
    \centering
    \begin{subfigure}{0.5\linewidth}
        \centering
        \includegraphics[width=.95\linewidth]{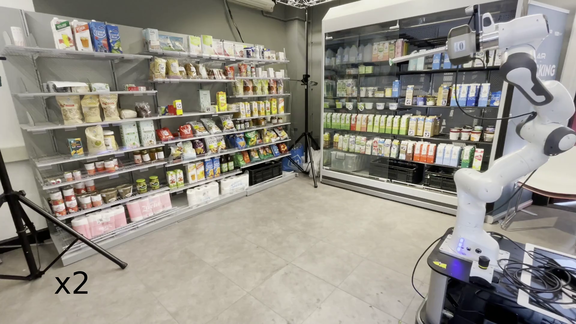}
        \label{fig:my_label}
    \end{subfigure}%
    \begin{subfigure}{0.5\linewidth}
        \centering
        \includegraphics[width=.95\linewidth]{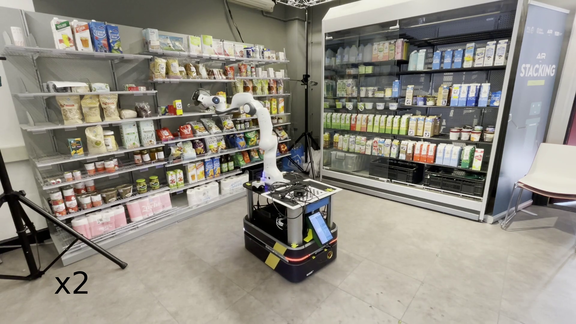}
        \label{fig:my_label}
    \end{subfigure}
    \par\smallskip
    \begin{subfigure}{0.5\linewidth}
        \centering
        \includegraphics[width=.95\linewidth]{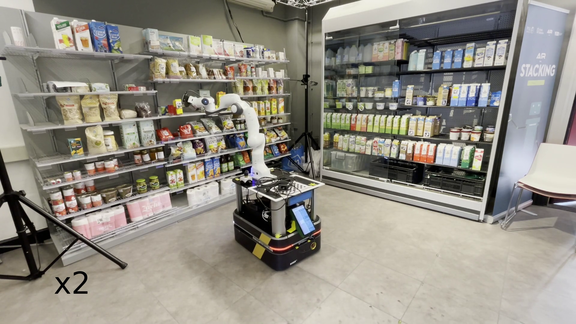}
        \label{fig:my_label}
    \end{subfigure}%
    \begin{subfigure}{0.5\linewidth}
        \centering
        \includegraphics[width=.95\linewidth]{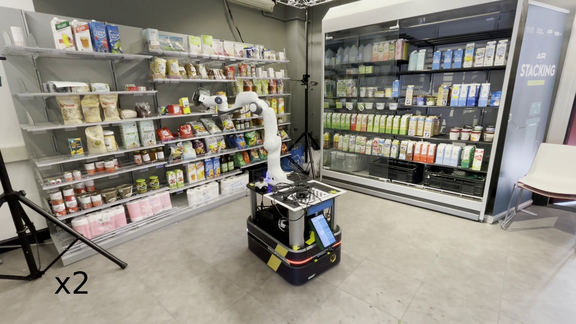}
        \label{fig:my_label}
    \end{subfigure}%
    \captionsetup{belowskip=-10pt}
    \caption{Trajectory generation with optimization fabrics for mobile manipulator using visual serving for product picking.}
    \vspace{-10pt}
\end{figure}
\section{Conclusion}%
\label{sec:conclusion}
We formulated parameter tuning for trajectory generation as a constrained
optimization problem. Additionally, we introduced symbolic optimization fabrics
that implement optimization fabrics in a parameterized way, for which the
general structure is pre-solved. The trajectory generator obtained with this
technique is parameterized and achieves low computational costs at runtime. We
showed that parameter tuning for symbolic optimization fabrics can be
effectively solved using Bayesian optimization. Additionally, we have shown
that the tuning generalized across different robots, tasks, and between
simulation and the real world. Finally, we qualitatively demonstrated that the
method applies to mobile manipulators. 
\changed[reviewer2]{
  While we aim at developing a method-agnostic autotuning framework for motion
  generation, symbolic optimization fabrics were selected as an example in 
  this work.
}

%


\newpage
\bibliographystyle{IEEEtran}
\bibliography{./src/references}

\end{document}